\pdfoutput=1

\documentclass[11pt]{article}

\usepackage[preprint]{acl}

\usepackage{times}
\usepackage{latexsym}

\usepackage[T1]{fontenc}

\usepackage[utf8]{inputenc}
\usepackage{hhline}
\usepackage{tabularx}
\usepackage{multirow}
\usepackage{adjustbox,booktabs}
\usepackage{mathtools}
\usepackage{footnote}
\usepackage{amssymb}
\usepackage{physics}

\usepackage{microtype}

\usepackage{inconsolata}

\usepackage{graphicx}
\usepackage{xspace}
\usepackage{algorithm}
\usepackage{makecell}
\newcommand{\ex}[1]{\textit{#1}\xspace}
\newcommand{\secref}[2][]{Section#1~\ref{#2}\xspace}
\newcommand{\figref}[2][]{Figure#1~\ref{#2}\xspace}
\newcommand{\tabref}[2][]{Table#1~\ref{#2}\xspace}
\newcommand{\eqnref}[2][]{Eqn#1.~(\ref{#2})\xspace}

\title{Static Word Embeddings for Sentence Semantic Representation}

\author{Takashi Wada
\, Yuki Hirakawa
\, Ryotaro Shimizu
\, Takahiro Kawashima
\, Yuki Saito\\
ZOZO Research \\
    [0ex]
    \texttt{\{firstname.lastname\}@zozo.com} 
    }

\begin{document}
\maketitle
\begin{abstract}

We propose new static word embeddings optimised for sentence semantic representation. We first extract word embeddings from a pre-trained Sentence Transformer, and improve them with sentence-level principal component analysis, followed by either knowledge distillation or contrastive learning. During inference, we represent sentences by simply averaging word embeddings, which requires little computational cost. We evaluate models on both monolingual and cross-lingual tasks and show that our model substantially outperforms existing static models on sentence semantic tasks, and even surpasses a basic Sentence Transformer model (SimCSE) on a text embedding benchmark. Lastly, we perform a variety of analyses and show that our method successfully removes word embedding components that are not highly relevant to sentence semantics, and adjusts the vector norms based on the influence of words on sentence semantics.
\end{abstract}

\section{Introduction}
In natural language processing (NLP), it has been a major topic to learn fixed-length embeddings that represent sentence semantics. To achieve this goal, most existing methods fine-tune pre-trained language models (LMs) such as BERT \cite{bert} using labelled data of various NLP tasks (e.g.\ natural language inference). Recent work further employs large language models (LLMs) such as LLaMA 3 \cite{dubey2024llama3herdmodels} as backbone models and shows that scaling up models can yield better performance \cite{behnamghader2024llmvec,lee2025nvembed}. However, this approach requires a large amount of computational cost, making it difficult to process billions of sentences cost-efficiently or deploy models on resource-constrained devices such as smartphones. As such, it is crucial to explore more cost-efficient models that work well without GPUs, and yet this direction remains largely overlooked in the research community.

In this work, we propose new static word embeddings (\textbf{SWEs}) --- both monolingual and cross-lingual ones --- that can represent sentence semantics fairly well with little computational cost. Specifically, our method first extracts SWEs from a pre-trained Sentence Transformer \cite{transformer,sbert}, and improves them with \textit{sentence-level} principal component analysis (PCA), followed by either knowledge distillation or contrastive learning. During inference, we encode sentences by simply looking up SWEs and taking their average, which can be done comfortably with limited resources. Our main contributions are: (1) we propose new SWEs that substantially outperform existing SWEs on a semantic textual similarity (STS) task, and even surpass SimCSE \cite{simcse} on Massive Text Embedding Benchmark \cite{mteb} in English; (2) we propose {cross-lingual} SWEs optimised for sentence representation, which achieve surprisingly good performance on translation retrieval even between distant languages (e.g.\ 94.6\% in F1 scores on English--Chinese); 
and (3) we conduct a variety of analyses and provide new insights into how to obtain SWEs suitable for sentence representation. We will release our code and models at \url{https://github.com/twadada/swe4semantics}.

\section{Related Work}

Most recent text embedding models are built on pre-trained Transformer LMs. For instance, GTE \cite{gte} fine-tunes BERT with contrastive learning on labelled data retrieved from various sources (e.g.\ STS data sets). Later, it is extended to the multilingual model called mGTE \cite{mgte}, which is trained on both monolingual and cross-lingual data sets. LLM2Vec \cite{behnamghader2024llmvec} scales up the backbone model and fine-tunes LLMs such as LLaMA with contrastive learning. Similarly, NV-Embed \cite{lee2025nvembed} fine-tunes Mistral-7B \cite{jiang2023mistral7b} with contrastive learning and also trains an additional attention layer on top. While these models achieve impressive results, they are computationally expensive and inefficient for real-world applications. To mitigate this, several methods to train compact models are proposed (e.g.\ knowledge distillation \cite{kd, xu-etal-2023-distillcse}), but they are mostly applied to Transformer models, which are designed with GPU acceleration in mind.

On the other hand, little attention has been paid to building cost-efficient models (esp.\ cross-lingual ones) that can be easily run on CPUs. Although such models inevitably lag behind heavy-weight models in accuracy, they offer significant advantages in cost and runtime efficiency, and can be applied to, for instance, quickly mining semantically similar content from billions of posts on social media. One traditional model in this direction is Sent2Vec \cite{pagliardini-etal-2018-unsupervised}, which extends the CBOW algorithm \cite{word2vec} and represents sentences by averaging static word and bigram embeddings. Very recently, although not published in papers, two static word embedding models called WordLlama \cite{wordllama} and Model2Vec \cite{model2vec} have been released from different GitHub projects (MIT license),\footnote{\url{https://github.com/dleemiller/WordLlama} and \url{https://github.com/MinishLab/model2vec}, respectively.} and demonstrate strong performance on sentence-level tasks in English. Based on their code and documentations, both models extract SWEs from Transformer models. Specifically, WordLlama extracts SWEs from the input layer of LLMs like LLaMa 3 \cite{dubey2024llama3herdmodels}, and refines them with contrastive learning on various labelled data such as STS and text retrieval data sets. Model2Vec extracts SWEs from a Transformer text encoder called BGE-base \cite{bge}, by feeding each word $w$ in a vocabulary into BGE without context (i.e.\ ``[CLS], $w$, [SEP]'') and using the output as the embedding of $w$. It then fine-tunes the SWEs by minimising the mean cosine distance between the sentence embeddings produced by BGE and the average of the SWEs for the same input text. To reduce the dimensionality, both WordLlama and Model2Vec apply PCA to a word embedding matrix as a post-processing step, and Model2Vec also applies the smooth inverse frequency (SIF) heuristic \cite{sif}, which multiplies the norm of $w$ by $\tfrac{\alpha}{\alpha+\mathrm{p}(w)}$, where $\mathrm{p}(w)$ denotes the uni-gram probability of $w$ and $\alpha=0.001$.

Amid the rapid growth in English models, there is a paucity of \textit{cross-lingual} SWEs optimised for sentence representation, as most existing ones are designed for word-level tasks such as bilingual lexicon induction and cross-lingual word alignment \cite{sup_map,xing-etal-2015-normalized,muse,artetxe-etal-2018-robust,wada-etal-2019-unsupervised,wada-etal-2021-learning}. Although Model2Vec also releases a multilingual model that extracts SWEs from LaBSE \cite{labse}, its cross-lingual performance is limited as we show in our experiments. We address this problem and propose effective SWEs not only for monolingual tasks but also for cross-lingual ones, such as (sentence) translation retrieval.

\section{Methodology} 

Our proposed method obtains SWEs in three steps: (1) embedding extraction from Sentence Transformer (Sec.\ \ref{sec_decontext}); (2) dimensionality reduction with \textit{sentence-level} PCA (Sec.\ \ref{sec_pca}); and (3) embedding refinement with knowledge distillation or contrastive learning (Sec.\ \ref{sec_kd}). Note that only the last step involves the training (fine-tuning) of SWEs.


\subsection{Embedding Extraction}\label{sec_decontext}
We first extract SWEs from a pre-trained Sentence Transformer (\textbf{ST}). Specifically, to obtain the embedding of a word $w \in V$ (denoted as $\mathrm{E}(w)$), we sample $N = 100$ sentences $z^{(w)}_i~(i=1,2,...,N)$ that contain $w$ from a large corpus.\footnote{In our preliminary experiments, setting $N=100$ did not bring a significant improvement compared to $N=30$. Therefore, we did not experiment with larger values for $N$.} Then, we encode each sentence $z^{(w)}_i$ using ST, which produces $\vert z^{(w)}_i\vert$ vectors at the last Transformer layer ($\vert z^{(w)}_i\vert$ denotes the number of tokens in $z^{(w)}_i$). We then retrieve the vector at the position of $w$ (denoted as $\mathrm{ST}(w, z^{(w)}_i)$), and obtain $\mathrm{E}(w)$ as:
\begin{gather}
\mathrm{E}(w)= \frac{1}{N}\sum_{i=1}^N{\mathrm{ST}(w, z^{(w)}_i)}.\label{eqn_distil}
\end{gather}
This way, we \textit{decontextualise} the embeddings and obtain the SWE of $w$. Furthermore, we can also generate cross-lingual SWEs using multilingual ST, as it embeds different languages in the same space. For the vocabulary $V$, we include more words than those in the ST's original vocabulary, and if $w$ is tokenised into subwords, we average its subword embeddings to obtain $\mathrm{ST}(w,z^{(w)}_i)$ in \eqnref{eqn_distil}. 

The idea of extracting SWEs from Transformers has been explored in previous studies \cite{ethayarajh-2019-contextual,bommasani-etal-2020-interpreting,wada-etal-2022-unsupervised}, which show that SWEs extracted from masked LMs perform well on word-level tasks (e.g.\ lexical substitution). However, we find that these embeddings are not effective for sentence-level tasks (as we will show in our ablation studies in \secref{sec_ablation}), and hence we further refine them in the following steps.

\subsection{Sentence-level PCA} \label{sec_pca}

Next, we apply {sentence-level} PCA to the extracted SWEs in \secref{sec_decontext} to improve performance and reduce the embedding dimensionality. In previous work (including WordLlama and Model2Vec), the dimensionality reduction of SWEs is typically done by applying PCA on a word embedding matrix $Y \in \mathbb{R}^{\vert V \vert \times d}$, where $d$ denotes the embedding dimension; we denote this as \textit{word-level} PCA. On the other hand, we apply PCA to a \textit{sentence} embedding matrix (as explained below), which emphasises the variance of sentences rather than words and improves performance on sentence-level tasks.

Specifically, we randomly sample $M = 100\mathrm{k}$ sentences from those used in \secref{sec_decontext}, and represent each sentence by averaging SWEs. When training cross-lingual SWEs, we sample $M$ sentences from $L$ languages and concatenate them; in our experiments, we train bilingual SWEs (i.e.\ $L=2$) unless otherwise specified. Given the sentence embedding matrix $X \in \mathbb{R}^{LM\times d}$ (with $L=1$ denoting monolingual SWEs) and its mean vector $\bar{X} \in \mathbb{R}^{d}$, we apply PCA to the centred matrix $X-\mathbf{1}_M \bar{X}^\top$, where $\mathbf{1}_M \in \mathbb{R}^{M }$ is a vector of ones, and then obtain the transformation matrix $W$. This matrix transforms the embedding space onto new axes that effectively capture the variance of the sentence representations. Importantly, since we represent each sentence by averaging SWEs, we can \textit{pre-transform} the SWE $\mathrm{E}(w)$ into $\mathrm{\hat{E}}(w)=W^\top(\mathrm{E}(w)-\bar{X})$ for each word $w \in V$ once and save $\mathrm{\hat{E}}(w)$ in memory, as $ W^\top(\frac{1}{\vert z\vert}\sum_{w'\in z}\mathrm{E}(w')-\bar{X}) = \frac{1}{\vert z\vert}\sum_{w'\in z}\mathrm{\hat{E}}(w')$.

To reduce the dimensionality using PCA, a common approach is to keep the principal components (PCs) with the $d' (<d)$ largest eigenvalues. However, we instead  discard the \textit{first}~$r$ PCs, and keep the $(r+1)$-th to $(r+d')$-th components. This is based on the previous finding that removing a few dominant PCs from SWEs can \textit{counter-intuitively} improve the embedding quality \cite{abtt}, and we actually get positive results in our experiments, especially in cross-lingual tasks (e.g.\ +49.4\% in F1 scores). Intriguingly, our analyses in \secref{sec_pca_analysis} reveal that it helps remove information that is not very relevant to sentence semantics. Following \citet{abtt}, we set $r$ to $\lfloor \frac{d}{100} \rfloor$, and refer to this method as \textbf{ABTT} (All-But-The-Top).

\subsection{Embedding Refinement}\label{sec_kd}
Lastly, we fine-tune the dimensionality-reduced embeddings obtained in \secref{sec_pca} to further optimise them for sentence representation. For monolingual SWEs, we perform teacher-student knowledge distillation, where we fine-tune our SWEs (= student) to reproduce the sentence similarities calculated by ST (= teacher). The teacher model is frozen during distillation and only SWEs are fine-tuned; therefore, it works with  modest computational resources. Besides, this refinement step does not require labelled data, unlike WordLlama. 

Formally, for a mini-batch of $K=128$ sentences, we minimise the following loss $\mathcal{L}$:
\begin{gather}
\mathcal{L} = -\frac{1}{K}\sum_{i=1}^{K}\sum_{j\neq i}^{K} \sigma\!\left({\mathbf{T},i,j}\right) \log\sigma\!\left({\mathbf{S},i,j}\right),\label{kd_loss}\\
\sigma(\mathbf{X},i,j) = \frac{e^{\frac{x_{i,j}}{\tau}}}{\sum_{k\neq i}^{K} e^{\frac{x_{i,k}}{\tau}}},\label{kd_sigma}
\end{gather}
where  $\mathbf{T}, \mathbf{S} \in \mathbb{R}^{K\times K}$ denote the similarity matrices calculated by the teacher and student models, with the value at the $i$-th row and $j$-th column corresponding to the cosine similarity between the $i$-th and $j$-th sentences in the mini-batch. The temperature $\tau$ is set to $0.05$ as commonly done in contrastive learning \cite{simcse}.\footnote{We did not extensively tune this hyper-parameter.} Note that the student model represents each sentence by averaging SWEs, while the teacher model encodes them with Transformers and applies its default pooling method. Since the teacher and student models calculate $T$ and $S$ independently, their embedding dimensions can be different and that makes it possible to fine-tune the dimensionality-reduced embeddings with PCA. In contrast, Model2Vec minimises the cosine distance of embeddings output by ST and SWEs, requiring both models to share the same embedding space. Consequently, it applies (word-level) PCA \textit{after fine-tuning}, and then needs to apply the SIF heuristic to achieve optimal performance. In \secref{sec_norm_analysis}, we will show that our method yields a similar effect to SIF but in a more refined manner.



For cross-lingual (bilingual) SWEs, we find that performing knowledge distillation for each language leads to {worse} results on cross-lingual tasks, as it does not encourage SWEs to be mapped into the same cross-lingual space. Therefore, we instead make use of a parallel corpus as a source of knowledge, and fine-tune the cross-lingual SWEs in languages $\ell1$ and $\ell2$ by minimising the following loss  $\mathcal{L}_{CL}$:
\begin{gather}
\mathcal{L}_{CL}= -\frac{1}{K}\sum_{i=1}^K (\log\sigma'(\mathbf{U},i) + \log\sigma'(\mathbf{U}^\top,i)),\label{eqn_cl_loss}\\
\sigma'(\mathbf{X}, i)= \frac{e^{\frac{x_{i,i}}{\tau}}}{\sum_{k =1}^{K} e^{\frac{x_{i,k}}{\tau}}},\label{cl_softmax}
\end{gather}
where $\mathbf{U}\in \mathbb{R}^{K\times K}$ denotes the similarity matrix calculated by SWEs, and $u_{i,j}$ denotes the cosine similarity between the $i$-th sentence in $\ell1$ and $j$-th sentence in $\ell2$ in the mini-batch. The $i$-th sentences in $\ell1$ and $\ell2$ are a pair of translations, and we create mini-batches by sampling $K$ translation pairs from a parallel corpus.\footnote{We transpose $U$ in \eqnref{eqn_cl_loss} to calculate the denominator in \eqnref{cl_softmax} for each language.} 

\section{Experiments}

\subsection{Our Models}
We train our monolingual (English) SWEs using GTE-base \cite{gte}, which achieves strong performance on Massive Text Embedding Benchmark (MTEB) among BERT-sized models. We sample sentences from a large-scale unlabelled corpus (CC-100 \cite{wenzek-etal-2020-ccnet,conneau-etal-2020-unsupervised}) and use them in each step of our method. Since CC-100 contains paragraphs as well as sentences, we segment text into sentences using a sentence splitter\footnote{We use the one at \url{https://github.com/microsoft/BlingFire}.} and retrieve $N$ short sentences for each word $w \in V$.\footnote{We find that using long text for extracting and fine-tuning our SWEs results in worse performance, likely due to the gap of information accessible to the teacher and student models during knowledge distillation (i.e.\ word order information).} For the vocabulary $V$, we include the 150k most frequent words (case-insensitive) in CC-100. 
To generate cross-lingual SWEs, we use mGTE \cite{mgte}, the multilingual version of GTE trained on 75 languages. We train bilingual embeddings for three language pairs, namely English--German (en--de), English--Chinese (en--zh), and English--Japanese (en--ja). We sample short sentences from large-scale parallel corpora (CC-Matrix \cite{schwenk-etal-2021-ccmatrix}) and use them to extract SWEs in \secref{sec_decontext} and calculate the cross-lingual loss in \eqnref{eqn_cl_loss}. For en--ja, we also use JParaCrawl v3.0 \cite{morishita-etal-2020-jparacrawl,morishita-etal-2022-jparacrawl} to augment data. For $V$ in English and German, we choose the 150k most frequent words (case-sensitive) in the corresponding corpora, and 30k most frequent \textit{tokens} (which can be a subword or consist of multiple words) in Japanese and Chinese.\footnote{This is because mGTE's tokeniser segments sentences into tokens without performing word segmentation in both languages (which have no whitespace word boundary).}

In \secref{sec_pca}, we reduce the dimensionality to $d' \in \{256,512\}$,\footnote{In most cases, 99\% of the variance can be explained with $d'=512$ when the original dimensionality $d$ is $768$.} and in  \secref{sec_kd} we fine-tune SWEs using Adam \cite{adam} with the learning rate of 0.001 for 30k steps, and apply early stopping based on the validation loss; the train and validation data consist of sentences sampled from those used in \secref{sec_decontext}. During inference, we tokenise the text $z$ into words (or tokens in Japanese and Chinese) and strip punctuation, and average SWEs to represent $z$. When a word is not in $V$, we tokenise it into subwords using (m)GTE's tokeniser, and truncate it by removing the last subword until the remaining string is found in $V$.\footnote{For instance, if \textit{tokeniser} is out of vocabulary, we split it into subwords (e.g.\ \textit{token \#ise \#r}), and strip the last subword(s). Then, if \textit{tokenise} (or \textit{token}) is in $V$, we use its embedding.}

\begin{table*}[t!]
\begin{center}
\begin{adjustbox}{max width=0.99\textwidth}
\begin{tabular}{llllllllllllll}
\toprule

  \multirow{1}{*}{Models ($d$)}&STS12&STS13&STS14&STS15&STS16&STS17&STS22&STS-B&SICK-R&BIO\\

  \midrule  
  \multicolumn{11}{c}{Sentence Transformers (STs)}\\\midrule
 MiniLM-L6 (384)&72.4&80.6&75.6&85.4&79.0&87.6&67.2&82.0&77.6&81.6\\
  SimCSE (768)&75.3&84.7&80.2&85.4&80.8&89.4&62.0&84.2&80.8&68.4\\
  BGE-base (768)&78.0&84.2&82.3&88.0&85.5&86.4&65.9&86.4&80.3&\textbf{86.9}\\

  GTE-base (768)&74.4&84.7&80.1&87.2&85.0&\textbf{90.6}&\textbf{68.6}&86.0&79.4&83.6\\
  LLM2Vec (4,096)&\textbf{79.3}&\textbf{84.8}&\textbf{82.9}&\textbf{88.1}&\textbf{86.5}&89.6&67.7&\textbf{88.0}&\textbf{83.9}&{84.9}\\\midrule
  \multicolumn{11}{c}{Static Word Embeddings (SWEs)}\\\midrule
  fastText (300)&57.2&69.2&62.8&73.0&64.2&70.2&52.1&56.5&60.1&63.0\\  
  Sent2Vec (700)&54.1&66.3&65.8&78.0&70.6&82.2&54.5&68.0&63.2&55.2\\  
  WordLlama (512)&63.9&73.3&69.1&81.5&76.1&85.2&60.2&77.0&67.0&71.3\\   
  Model2Vec (512)&62.7&77.6&72.9&80.8&76.9&\textbf{87.1}&\textbf{64.3}&76.8&65.7&77.6\\   
  \textbf{OURS (256)}&\textbf{68.3}&\textbf{79.3}&\textbf{75.9}&\textbf{83.1}&\textbf{79.4}&86.1&59.3&\textbf{79.2}&\textbf{68.0}&\textbf{80.3}\\   
  
\bottomrule
\end{tabular}
\end{adjustbox}
\caption{Results on STS data sets. The best scores among the ST and SWE models are boldfaced.} \label{sts_results}
\end{center}
\end{table*}

\subsection{Baselines}

In monolingual experiments, we include two traditional SWEs as our baselines, namely fastText \cite{fasttext} and Sent2vec. Both models are self-supervised based on word co-occurrence information, with Sent2vec performing better on sentence-level tasks. We also compare our model against two recent GitHub-hosted models: WordLlama and Model2Vec. We use \textit{wordllama-l3-supercat} for WordLlama, which is based on LLaMa 3, and \textit{potion-base-32M} for Model2Vec, the latest and state-of-the-art model (released in January 2025) based on BGE-base. SWE models, including ours, produce sentence vectors by averaging SWEs and applying L2-normalisation.

In cross-lingual experiments, we compare our model against the multilingual Model2Vec model (\textit{M2V\_multilingual\_output}). For English--German, we also evaluate the cross-lingual SWE available at the MUSE GitHub repository\footnote{\url{https://github.com/facebookresearch/MUSE}}~\cite{muse}, which aligns English and German fastText embeddings using a bilingual dictionary. This model has been widely used in research papers.

\begin{table}[t!]
\begin{center}
\begin{adjustbox}{max width=0.99\columnwidth}
\begin{tabular}{llllll}
\toprule

  \multirow{2}{*}{Models ($d$)}&\multirow{2}{*}{$100\rho$}&\multicolumn{2}{c}{runtime (s)}\\\cmidrule{3-4}
    &&\multicolumn{1}{c}{GPU}&\multicolumn{1}{c}{CPU}\\

  \midrule  
  MiniLM-L6 (384)&85.4&1.7&8.8\\   
  GTE-base (768)&87.2&3.4&52.7\\
  LLM2Vec (4,096)&88.1&30.3&$>$10,000\\   \midrule    
  WordLlama (512)&81.5&--&0.4\\   
  Model2Vec (512)&80.8&--&0.3\\   
  \textbf{OURS (256)}&83.1&--&0.4\\

\bottomrule
\end{tabular}
\end{adjustbox}
\caption{Scores and runtime (in second) on STS15. The runtime is averaged over three independent runs, and is measured on the same virtual machine with or without enabling a GPU (NVIDIA A100-SXM4-40GB).} \label{speed_comparison}
\end{center}
\end{table}

\subsection{Experiments on English Tasks}
We first evaluate monolingual (English) models on semantic textual similarity (STS), a well-established NLP task for sentence semantic evaluation. It is a task of calculating textual similarity for each sentence pair, and models are evaluated based on Spearman's rank correlation $\rho$ between their predictions and human judgements. \tabref{sts_results} shows the results on STS12--17, 22, STS Benchmark (STS-B), SICK-R, and BIOSSES (BIO) data sets~\cite{sts12,sts13,sts14,sts15,sts16,sts22,sick-r,BIOSSES}. The first five rows show performance of STs for reference (\textit{not} as baselines of our model): SimCSE\footnote{We use the supervised one, which performs better than the unsupervised one.} is a pioneering ST, which often serves as a strong baseline in STS research; MiniLM-L6\footnote{sentence-transformers/all-MiniLM-L6-v2} is a popular light-weight ST based on MiniLM \cite{minilm}; BGE and GTE are the models used to train Model2Vec and OURS, respectively; and LLM2Vec is a recent large-scale model built on LLaMa-3-8B. The table shows that OURS substantially outperforms all SME baselines on most data sets, even though it has the smallest dimensionality ($d=256$). Our model also outperforms MiniLM-L6 on STS14 and STS16 and SimCSE on BIOSSES, where the sentences are sampled from biomedical domains, indicating the robustness of OURS to various types of text. \tabref{speed_comparison} compares the scores and runtime on STS15. While falling behind STs in accuracy, OURS (and other SWEs) can run extremely fast (20 times faster than MiniLM on CPUs) without sacrificing the accuracy very much.

Additionally, to further evaluate the effectiveness of our model on other tasks such as sentence classification, we also run experiments on Massive Text Embedding Benchmark (MTEB) \cite{mteb}. It consists of 56 data sets covering 7 different tasks, namely classification, pair classification, clustering, reranking, retrieval, STS, and summarisation evaluation. On each task, a model encodes text and returns its fixed-length vector, which is then used on downstream tasks (e.g.\ for calculating textual features or similarities). Models are evaluated using task-specific metrics (e.g.\ precision) on each data set. Since MTEB contains both sentence-level and document-level tasks, and our focus is on sentence semantic representation, we report two scores: \textbf{Avg}, the average across all 56 data sets; and \textbf{Avg-s2s}, the average across the 33 data sets labelled as ``s2s'', where the input text is a {sentence} rather than a document. \tabref{mteb_ave_results} presents the results, showing that our model substantially outperforms the other SWEs both in Avg and Avg-s2s, even with the smallest embedding size. Moreover, our model outperforms SimCSE, which is surprising given that our model does not use any sequential function like Transformer. Interestingly, we also find that ensembling OURS with WordLlama and Model2Vec further boosts performance (64.55 and 53.28 in Avg-s2s and Avg), which we describe in detail in Appendix \ref{sec_ensembling}. We also show the scores on each task in \tabref{result_table} in Appendix. 

\begin{table}[t!]
\begin{center}
\begin{adjustbox}{max width=0.99\columnwidth}
\begin{tabular}{lllllllll}
\toprule

  \multirow{1}{*}{Models ($d$)}&Avg-s2s&Avg\\

  \midrule  
  MiniLM-L6 (384)&65.95&56.26\\   
  SimCSE (768)&62.75&48.86\\    
  BGE-base (768)&71.18&63.55\\   
  GTE-base (768&71.51&64.11\\
  LLM2Vec (4,096)&\textbf{72.94}&\textbf{65.01}\\\midrule
  fastText (300)&53.68&43.05\\  
  Sent2Vec (700)&56.33&46.29\\  
  WordLlama (512)&60.40&50.23\\   
  Model2Vec (512)&62.37&51.33\\   
  \textbf{OURS (256)}&63.76&51.97\\   
  \textbf{OURS (512)}&\textbf{64.06}&\textbf{52.35}\\
  
\bottomrule
\end{tabular}
\end{adjustbox}
\caption{Results on MTEB experiments in English. The best scores among STs and SWEs are bold-faced. Avg of Model2Vec is slightly lower the reported one due to a small bug in their MS MARCO evaluation.} \label{mteb_ave_results}
\end{center}
\end{table}

\begin{table}
\begin{center}
\begin{adjustbox}{max width=\columnwidth}
\begin{tabular}{ll@{\;}@{\;}ll@{\;}@{\;}l@{\;}@{\;}ll}
\toprule

  \multirow{2}{*}{Models ($d$)}&\multicolumn{2}{c}{BUCC}&\multicolumn{3}{c}{Tatoeba}\\
  \cmidrule(lr){2-3} \cmidrule(lr){4-6}
     &en-de&en-zh&en-de&en-zh&en-ja\\       
  \midrule  
  mGTE (768)&98.6&98.2&97.8&95.7&93.1\\\midrule   
  MUSE (300)&46.6&--&44.3&--&--\\   
  Model2Vec (256)&60.0&0.0&73.5&26.6&15.3\\   
  OURS (256) &\textbf{96.3}&\textbf{94.6}&\textbf{95.1}&\textbf{87.9}&\textbf{79.0}\\

\bottomrule
\end{tabular}
\end{adjustbox}
\caption{Results (F1) on translation retrieval tasks.} \label{tab_translation}
\end{center}
\end{table}
\begin{table}[t]
\begin{center}
\begin{adjustbox}{max width=0.99\columnwidth}
\begin{tabular}{lll@{\;}@{\;}l@{\;}@{\;}l@{\;}@{\;}@{\;}ll}
\toprule
\multirow{2}{*}{Models ($d$)}&\multicolumn{1}{c}{STS17}&\multicolumn{2}{c}{STS22}\\
\cmidrule(lr){3-4}
  &en-de&en-de&en-zh\\
  \midrule 
  mGTE (768)&84.7&62.3&72.9\\
  \midrule   
  MUSE (300)&11.2&36.9&--\\
  Model2Vec (256)&54.2&39.3&26.2\\
  OURS (256) &\textbf{70.5}&\textbf{62.0}&\textbf{72.6}\\

\bottomrule
\end{tabular}
\end{adjustbox}
\caption{Results on cross-lingual and monolingual (German and Chinese) STS tasks. } \label{xlingal_sts}
\end{center}
\end{table}

\subsection{Experiments on Cross-Lingual Tasks} \label{xlingual_results}

We first evaluate cross-lingual SWEs on translation retrieval, a task of retrieving the translation of the input text based on cosine similarity. \tabref{tab_translation} shows the results (F1 scores) on BUCC \cite{bucc} and Tatoeba \cite{artetxe-schwenk-2019-massively,tatoeba} data sets for three language pairs: English--\{German, Chinese, Japanese\}. The first row shows the performance of mGTE, and the rest are cross-lingual SWEs. The table clearly shows that OURS outperforms the SWE baselines by a large margin, and performs fairly well on all language pairs (e.g.\ 96.3\% for English-German on BUCC). This suggests the applicability of our model as a light-weight approach to extracting translation candidates, which could be further validated by computationally expensive models such as mGTE. Lastly, we also evaluate SWEs on cross-lingual STS (STS17 and STS22), a task of measuring textual similarity across two different languages, and \tabref{xlingal_sts} presents the results. Again, OURS is the best performing SWE, demonstrating its effectiveness on cross-lingual tasks. 

It should be noted that mGTE and Model2vec are multilingual models (i.e.\ aligned across more than two languages) whereas MUSE and OURS are bilingual ones, and this can place our model in a favourable position in bilingual evaluation. Given this concern, we also try generating multilingual SWEs aligned across the four languages (English, German, Chinese, and Japanese) using our model, and find that it still outperforms the SWE baselines by a large margin; see Appendix \ref{multilingual_embs} for details.

\section{Analysis}

\subsection{Ablation Studies}\label{sec_ablation}

\begin{table}[t!]
\begin{center}
\begin{adjustbox}{max width=0.98\columnwidth}
\begin{tabular}{llllll}
\toprule

  &en&&en-de&en-zh&en-ja\\
  \midrule  
  \secref{sec_decontext}& 44.2&& 51.6& 45.0&28.6 \\   
  \secref{sec_pca}&49.9&&93.8& 88.6& 74.6\\   
  ~~~w/o ABTT&48.8&& 49.6& 39.2& 33.2\\

  \secref{sec_kd}&\textbf{52.0}&& \textbf{95.7}& \textbf{91.2}& \textbf{79.0}\\   
  
\bottomrule
\end{tabular}
\end{adjustbox}
\caption{Performance of our SWEs obtained at each step. The ``en'' column denotes Avg on MTEB, and the rest are the average scores on BUCC and Tatoeba. The third row shows the results without ABTT in \secref{sec_pca}.} \label{tab_ablation}
\end{center}
\end{table}

To verify the effectiveness of our method, we evaluate our SWEs ($d=256$) obtained in Sections \ref{sec_decontext}, \ref{sec_pca}, and \ref{sec_kd}, and see whether the performance improves after each step. \tabref{tab_ablation} shows the results. The column under ``en'' shows performance on MTEB (Avg), demonstrating that sentence-level PCA and knowledge distillation substantially improve the SWEs extracted in \secref{sec_decontext}. Importantly, our model surpasses the performance of fastText (43.1) \textit{even without fine-tuning in Section~\ref{sec_kd}\xspace}, demonstrating the effectiveness of extracting SWEs from Sentence Transformers. Other important findings that are \textbf{not} presented in \tabref{tab_ablation} (due to space restrictions) include: (1) applying \textit{word-level} PCA in \secref{sec_pca} performs worse than \textit{sentence-level} PCA (48.1 vs.\ 49.9); and (2) performing knowledge distillation without PCA (i.e.\ ablating \secref{sec_pca}) results in poor performance (44.4)\footnote{Performance increases on some tasks but drops on others.}, indicating that parameter initialisation is a key to success in knowledge distillation. The last three columns in \tabref{tab_ablation} show the average scores on BUCC and Tatoeba for each language pair. The results indicate that the SWEs obtained in \secref{sec_pca} already achieve good scores, and fine-tuning them in \secref{sec_kd} further boosts performance. Another important observation is that applying ABTT in \secref{sec_pca}, i.e.\ removing the \textit{first} $r$ PCs from SWEs, plays a vital role in generating cross-lingual SWEs; we will provide an in-depth analysis on its effect in \secref{sec_pca_analysis}.

\tabref{tab_diffrent_STs} shows the performance of our models trained with different STs, namely SimCSE, Nomic \cite{nussbaum2024nomic},\footnote{Nomic is trained to encode text with a task-specific prefix prompt, and hence we prepend the prompt "classification: " to every input text in \secref{sec_decontext} and \secref{sec_kd}.} and GTE-\{small/base/large\}. It indicates that OURS performs better using better STs (SimCSE < Nomic < GTE) as its teacher model,\footnote{Note that SimCSE uses $[$CLS$]$ pooling while Nomic and GTE use mean pooling for generating sentence embeddings. While our model works reasonably well with both model types, this could be another reason why it performs better with Nomic and GTE than SimCSE, as our method extracts SWEs in \secref{sec_decontext} by averaging embeddings across $N$ sentences.} suggesting that its performance might further increase as ST models evolve. That being said, our model does not benefit from using the largest GTE model, possibly due to the large discrepancy of the parameter sizes between the teacher and student models during knowledge distillation. Performance on each MTEB task is shown in \tabref{tab_other_model} in Appendix.

\begin{table}[t!]
\begin{center}
\begin{adjustbox}{max width=0.98\columnwidth}
\begin{tabular}{lllllll}
\toprule
   &SimCSE&Nomic&\multicolumn{3}{c}{GTE}\\
  \cmidrule{4-6}
  $d$&768&768&384&768&1,024\\
  \midrule  
  ST&62.8&69.5&69.3&{71.5} &\textbf{71.7}\\   
  OURS&61.2&62.7&60.8 &\textbf{63.8}&62.9\\     
\bottomrule
\end{tabular}
\end{adjustbox}
\caption{Performance (Avg-s2s on MTEB) of different STs and OURS (256) trained with them.} \label{tab_diffrent_STs}
\end{center}
\end{table}

\subsection{Effects of PCA}\label{sec_pca_analysis}

\begin{figure*}[t] 
    \centering
    \includegraphics[width=\textwidth]{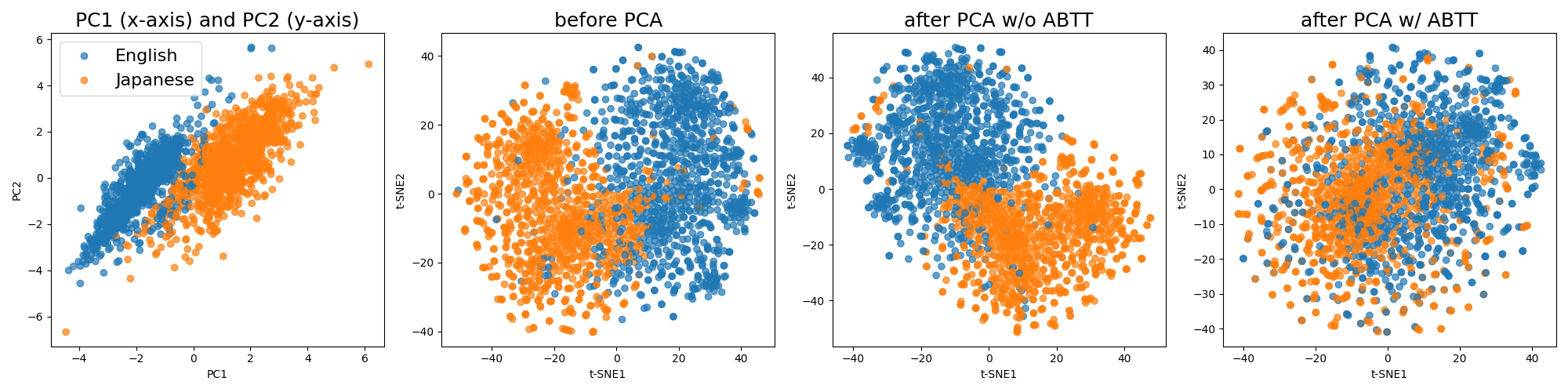} 
    \caption{Scatter plots of our cross-lingual SWEs in English (blue) and Japanese (orange). The leftmost shows the 1st and 2nd principal components, and the rest are t-SNE visualisation before and after applying PCA wo/w ABTT.}
    \label{fig:tsne-enja}
\end{figure*}

 To investigate why PCA with ABTT is particularly effective on cross-lingual tasks, we analyse what information is encoded in the first two PCs, and find that it captures the variance largely based on the \textit{language identity}. To visualise this, we randomly sample 1,000 language-specific words/tokens from English and Japanese and plot their embeddings based on the values in the first two PCs (PC1 and PC2). The result is presented in \figref{fig:tsne-enja} (the leftmost plot), which clearly shows that the embeddings are separated based on language. Given this finding, we also visualise the embeddings using t-SNE \cite{tsne} before and after applying PCA wo/w ABTT, and the results are also shown in \figref{fig:tsne-enja}. We can see that the embeddings create language clusters before PCA and after PCA \textit{without} ABTT, but applying PCA \textit{with} ABTT renders the embeddings more language-agnostic by removing the language-identity components.\footnote{We also confirm this finding by measuring the silhouette scores, which are: 0.10 before PCA; 0.11 after PCA w/o ABTT; and 0.02 after PCA w/ ABTT, respectively.} We also observe the same trend in English--Chinese/German, as shown in Figures \ref{fig:tsne-enzh} and \ref{fig:tsne-ende} in Appendix. Such language-specific dimensions can act as noise when measuring semantic similarity across different languages,\footnote{For instance, Japanese sentences with many English loanwords can have high similarities to English sentences regardless of the semantics.} explaining why ABTT is crucial in cross-lingual SWEs. Furthermore, this observation makes us wonder whether such language segregation also occurs in the mGTE embedding space; we investigate this and find that it \textit{does} create language clusters at lower layers of Transformer, but generates more language-agnostic embeddings at higher layers. Since it deviates from the main subject of this study, we discuss more details in Appendix \ref{mgte_analysis}.

\begin{table}[t]
\begin{center}
\begin{adjustbox}{max width=\columnwidth}
\begin{tabular}{lllc}
\toprule
$i$&\multicolumn{1}{c}{Smallest}&\multicolumn{1}{c}{Largest}\\
\midrule
1&\makecell[c]{advantages, maps, permits, \\categories, manufacturers}&\makecell[c]{haha, ha, grinned, \\omg, oh}\\\midrule
2&\makecell[c]{appreciate, hey, \\haha, btw, anyways}&\makecell[c]{shook, sheriff, snapped, \\glanced, sighed}
\\\midrule
6&\makecell[c]{nov, oct, dec, \\2020, feb}&\makecell[c]{charming, compliment, 
\\courteous, friendly, softly}
\\
\bottomrule
\end{tabular}
\end{adjustbox}
\end{center}
\caption{Examples of words with the smallest and largest values in the $i$-th principal component.} 
\label{pca_removal_examles}
\end{table}

Lastly, we also analyse the dominant PCs of monolingual (English) SWEs, and find that some of them seem to capture linguistic styles or domains. Table \ref{pca_removal_examles} shows examples of the words that have the five largest and smallest values in PC1, PC2, and PC6 among the 10k most frequent words in $V$. In PC1 and PC2, words seem to be clustered based on the word frequency in casual text, whereas in PC6, date-related words seem to have large values. To verify whether PC1 and PC2 are actually related to textual formality, we conduct a simple experiment using the data set released by \citet{chhaya-etal-2018-frustrated}, where 10 human annotators judge the formality level of 960 emails. We encode each email by averaging SWEs and measure the Pearson correlation coefficient between the values in PC1, PC2, or PC6, and the mean scores of the 10 annotators. We observe that the correlations are: $-0.63$, $0.37$, and $-0.11$ for PC1, PC2, and PC6, confirming a moderate or weak correlation for the first two PCs.\footnote{We surmise that this may be attributed to the use of sentences sampled from a web corpus in our model, which is a mixed bag of formal and casual text. It would be interesting to see how the selection of text data affects the outcome.}  Albeit those components capture the variance of sentences effectively, they would not necessarily serve as useful features on semantic tasks, explaining why PCA w/ ABTT works well for monolingual SWEs as well.\footnote{But on classification tasks, ABTT often harms performance.}

\begin{figure}[t] 
    \centering
    \includegraphics[width=1\columnwidth]{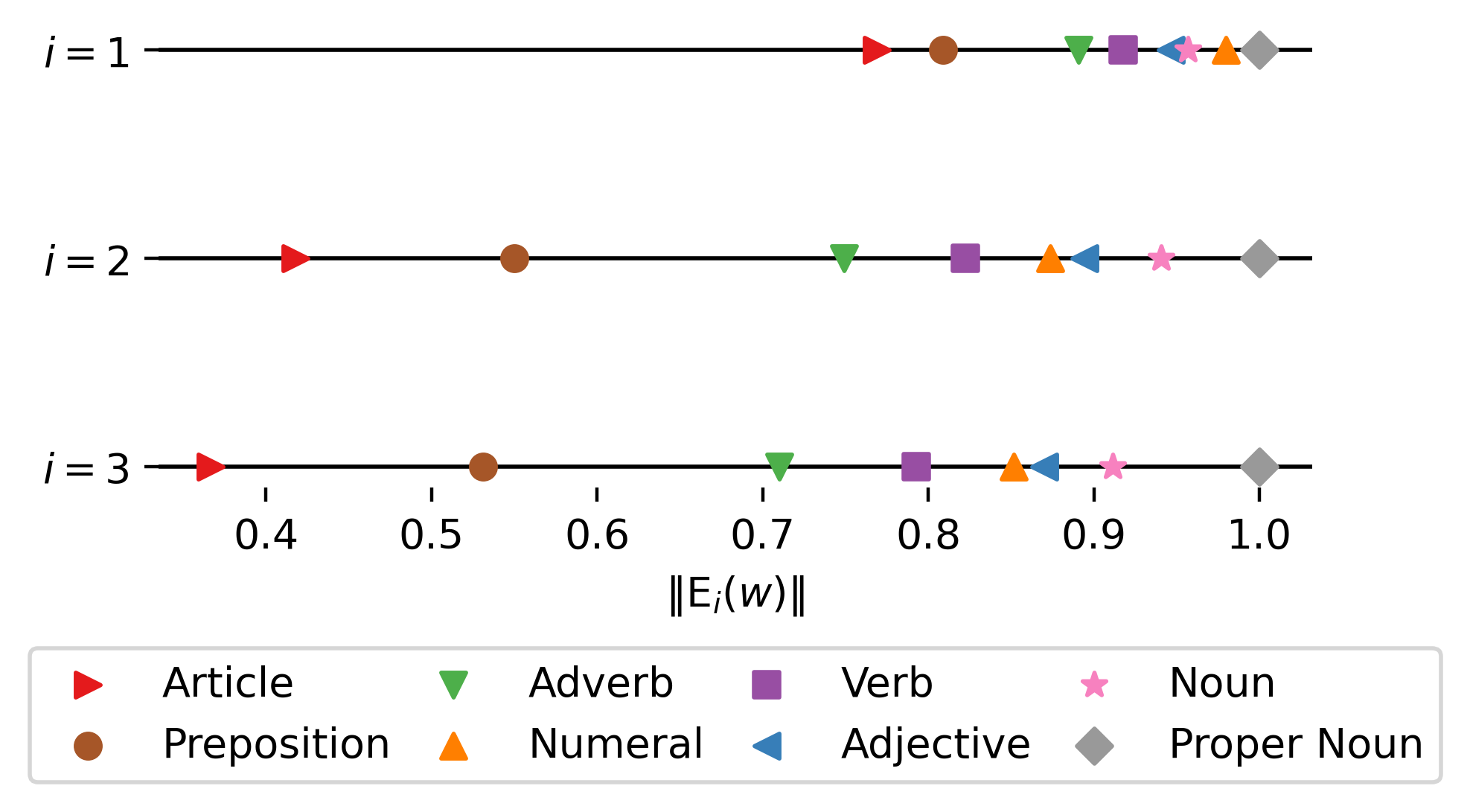} 
    \caption{Comparison of word embedding norms obtained in Sections \ref{sec_decontext}, \ref{sec_pca}, and \ref{sec_kd}. For normalisation, norms of each POS tag are divided by the maximum value of all tags (i.e.\ the norm of Proper Noun).}
    \label{fig:norms}
\end{figure}

\subsection{Analysis of Word Embedding Norm}\label{sec_norm_analysis}

To investigate how our method optimises word embeddings, we analyse the norms of SWEs obtained in Sections \ref{sec_decontext}, \ref{sec_pca}, and \ref{sec_kd}; we denote them as  $\|\mathrm{E}_i(w)\| $ ($i=1,2,3$). For the purpose of analysis, we compare the norms of the 10k most frequent words in Brown Corpus, which consists of text from various domains and provides part-of-speech (POS) tags for each word. First, we compare the mean norms of word embeddings for each POS tag; for words that are annotated with multiple POS tags (e.g.\ \ex{notice}), we use the most frequent one. The results are shown in \figref{fig:norms}. For normalisation, we divide the norms of each POS tag by that of ``Proper Noun'', which always has the largest value regardless of $i$. It shows that the differences in the norms are emphasised after PCA and knowledge distillation. In particular,  the relative norms of ``Article'' (e.g.\ \textit{the}), ``Preposition''   (e.g.\ \textit{of}), and ``Numeral''   (e.g.\ \textit{2025}, \textit{first}) drop sharply, likely because these types of words do not affect sentence semantics very much. On the other hand, ``Proper Noun'' and  ``Noun'' have the largest norms after PCA and knowledge distillation, as they often convey information relevant to semantics. This result is somewhat akin to the previous finding that ST tends to emphasise nominal information of a sentence \cite{nikolaev-pado-2023-representation}. In Appendix \ref{pos_analysis_detail}, we also perform the same analysis for SWE baselines (fastText, WordLlama, and Model2Vec) and reveal that fastText assigns relatively large norms to articles, suggesting it is not well optimised for sentence representation.

Lastly, we find that our method induces a similar effect to SIF, a heuristic used in Model2Vec --- we measure Spearman's rank correlation between the norm and word frequency rank and find that it increases after PCA and knowledge distillation ($0.39 \rightarrow 0.43 \rightarrow 0.49$), meaning both steps decrease the norms of high-frequency words. Compared to SIF, however, knowledge distillation adjusts norms in a more nuanced way, considering not only word frequency (which is implicitly encoded by ST \cite{kurita-etal-2023-contrastive}) but also POS tags and influence of each word on sentence semantics.

\subsection{Impact of Model Size}\label{effect_d_v_sec}  

\begin{table}
\begin{center}
\begin{adjustbox}{max width=1\columnwidth}
\begin{tabular}{llllll}
\toprule

  $d$&$\vert V \vert$&Avg-s2s&Avg\\
  \midrule  
  512&150k&\textbf{64.1}& \textbf{52.4}\\   
  256&150k&63.8& 52.0\\   
  128&150k&62.6&50.4\\   
  64&150k&52.5&41.4\\ \midrule  
  256&100k&63.5&51.8\\   
  256&75k&63.4&51.5\\   
  256&50k&62.7&51.0\\   
  256&30k&62.2&50.2\\     
  
\bottomrule
\end{tabular}
\end{adjustbox}
\caption{Performance of our model on MTEB with different embedding and vocabulary sizes.} \label{tab_vocab_dim_effect}
\end{center}
\end{table}

\tabref{tab_vocab_dim_effect} compares OURS trained with different dimensionalities $d$ and vocabulary sizes $\vert V \vert$. Our model performs the best with $d\geq256$ and also achieves good scores with $d=128$. However, the scores drop sharply with $d=64$, where knowledge distillation rather harms performance (44.0 $\rightarrow$ 41.4) likely due to the large discrepancy of the embedding spaces between the teacher and student models. In terms of $\vert V \vert$, our model performs well with $\vert V \vert  \geq 75\mathrm{k}$. Hence, setting $d=256$ and $\vert V\vert\geq75\mathrm{k}$ would offer a good balance between performance and memory usage (in English). We provide scores on each task in  \tabref{detail_model_size} in Appendix.

\section{Conclusion}

We propose new word embeddings optimised for sentence semantic representation. We extract embeddings from Sentence Transformers and improve them with sentence-level PCA followed by knowledge distillation or contrastive learning. Our rigorous experiments show that our model outperforms baselines on both monolingual and cross-lingual semantic tasks. Our in-depth analyses on word embeddings also reveal that our method successfully removes embedding components that are not very relevant to sentence semantics, and adjusts the vector norms based on the influence on sentence semantics.

\section{Limitations}
Since our model (and other SWEs) does not consider word order information (i.e.\ a bag-of-words model), its performance is limited on tasks where models are required to understand long context, such as document retrieval. Importantly, however, there is \textit{always} a trade-off between model performance and computational complexity; in this work, we focus on improving the cost-efficiency, an important factor for real-world applications that is nonetheless largely neglected among the research community these days. To improve performance (at the cost of computational cost), one can simply add a small network that can capture sequential information (e.g.\ Transformer, CNN, LSTM \cite{lstm}) on top of our word embeddings, and train them with labelled data as performed during the training process of Sentence-Transformer models. Another idea to better capture sentence semantics is to represent text with a set of word embeddings with positional information (rather than taking their average), and calculate textual similarity by measuring the distance (or optimal transport cost) between two distributions, e.g.\ Word Mover's Distance \cite{pmlr-v37-kusnerb15}.

\bibliography{acl_latex}
\appendix

\begin{table}[ht]
\begin{center}
\begin{adjustbox}{max width=0.98\columnwidth}
\begin{tabular}{llllll}
\toprule

 Models &en-de&en-zh&en-ja\\
  \midrule  
  {Bilingual}& 95.7& 91.2&79.0\\   
{Multilingual}& 93.1&86.6&71.6\\   
\bottomrule
\end{tabular}
\end{adjustbox}
\caption{Performance of our bilingual and multilingual SWEs on translation retrieval tasks. The values denote the average scores on BUCC and Tatoeba for each language pair. Note that the multilingual model is obtained \textit{without fine-tuning} in \secref{sec_kd}}. \label{multilingual_retrieval}
\end{center}
\end{table}

\begin{table}[ht]
\begin{center}
\begin{adjustbox}{max width=0.98\columnwidth}
\begin{tabular}{llllll}
\toprule
 \multirow{2}{*}{Models}&\multicolumn{1}{c}{STS17}&\multicolumn{2}{c}{STS22}\\
\cmidrule(lr){3-4}
  &en‑de&en‑de&en‑zh\\\midrule  
  {Bilingual}&70.5&62.0&72.6\\   
{Multilingual}&69.7&60.6&70.4\\   
\bottomrule
\end{tabular}
\end{adjustbox}
\caption{Performance of our bilingual and multilingual SWEs on cross-lingual STS. Note that the multilingual model is obtained \textit{without fine-tuning} in \secref{sec_kd}.} \label{multilingual_sts}
\end{center}
\end{table}

\section{Results on Multilingual Embeddings}\label{multilingual_embs}

In addition to bilingual SWEs, we also try generating multilingual SWEs using our model, where English, German, Chinese, and Japanese are all embedded in the same space (as in Model2Vec). To achieve this, we apply the sentence-level PCA  to the concatenation of the sentence vectors sampled from each language (i.e.\ setting $L$ to $4$ in \secref{sec_pca}). For multilingual SWEs, \textit{we skip the fine-tuning step in Section~\ref{sec_kd}\xspace} because it is specifically designed for bilingual training; however, note that it could be easily extended to multilingual training by jointly optimising the bilingual loss $\mathcal{L}_{CL}$ for multiple language pairs. \tabref{multilingual_retrieval} and \tabref{multilingual_sts} compare the performance of our bilingual and multilingual SWEs. Although the multilingual model performs worse than the bilingual one (which benefits from fine-tuning), it still substantially exceeds the performance of the SWE baselines (MUSE and Model2Vec) shown in \tabref{tab_translation} and \tabref{xlingal_sts}.

\section{Analysis of Baseline Embedding Norms}\label{pos_analysis_detail}

\begin{table}[t!]
\begin{center}
\begin{adjustbox}{max width=\columnwidth}
\begin{tabular}{lllllllll}
\toprule

  POS&OURS&FT&WL&M2V\\

  \midrule  
  Article&0.37&0.55&0.22&0.24\\   
  Preposition&0.53&0.66&0.33&0.57\\   
  Adverb&0.71&0.72&0.52&0.85\\   
  Numeral&0.85&0.76&0.66&0.79\\   
  Verb&0.79&0.81&0.67&0.86\\   
  Adjective&0.87&0.81&0.74&0.90\\  
  Noun&0.91&0.86&0.84&0.91\\
  Proper Noun&1.00&1.00&1.00&1.00   \\
  
\bottomrule
\end{tabular}
\end{adjustbox}
\caption{Comparison of word embedding norms generated by fastText (FT), WordLlama (WL), Model2Vec (M2V), and our model (OURS). For normalisation,
norms of each POS tag are divided by the maximum value of all tags, which is the norm of Proper Noun for all models.} \label{pos_analysis_baselines}
\end{center}
\end{table}

\tabref{pos_analysis_baselines} compares word embedding norms generated by fastText (FT), WordLlama (WL), Model2Vec (M2V), and our model (OURS). Each value is normalised (divided) by the norm of Proper Noun within each model, as done in Figure 2. Interestingly, it shows that Proper Noun has the largest norm for all models. Looking at each model, fastText has relatively large norms for Article and Preposition (which usually have a small influence on sentence semantics), suggesting it is not well optimised for sentence representation (as we demonstrated in our experiments). For WordLlama, only  Noun and Proper Noun have larger values than 0.8, suggesting it emphasises nominal information more than the other models. This partially explains why it performs well on document retrieval tasks on MTEB (as shown in \tabref{result_table}), where capturing topical information is arguably more important than the semantics. Model2Vec and OURS display similar patterns, with Model2Vec assigning a higher weight to Verb and Adverb and OURS emphasising Numeral more.

\section{Analysis of mGTE Embeddings}\label{mgte_analysis}

\begin{figure*}[t] 
    \centering
    \includegraphics[width=\textwidth]{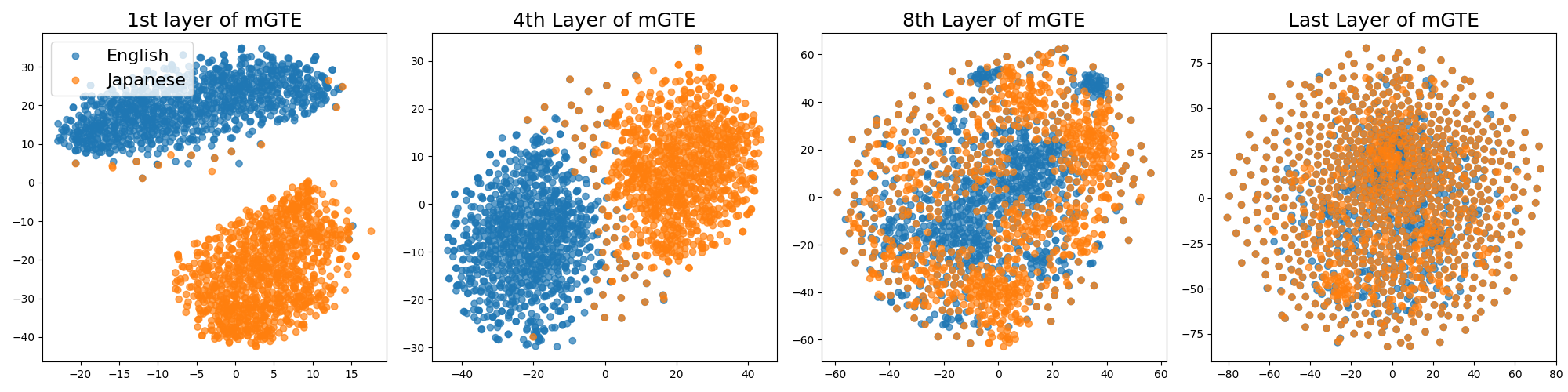} 
    \caption{The t-SNE visualisation of sentence embeddings produced by \textbf{mGTE-base} at the 1st, 4th, 8th, and 12th (= last) layers of Transformer for English (blue) and Japanese (orange) sentences. The encoded sentences are 1,000 pairs of translations sampled from a parallel corpus.}
    \label{fig:tsne-gte-layers}
\end{figure*}

\begin{figure*}[t!] 
    \centering
    \includegraphics[width=\textwidth]{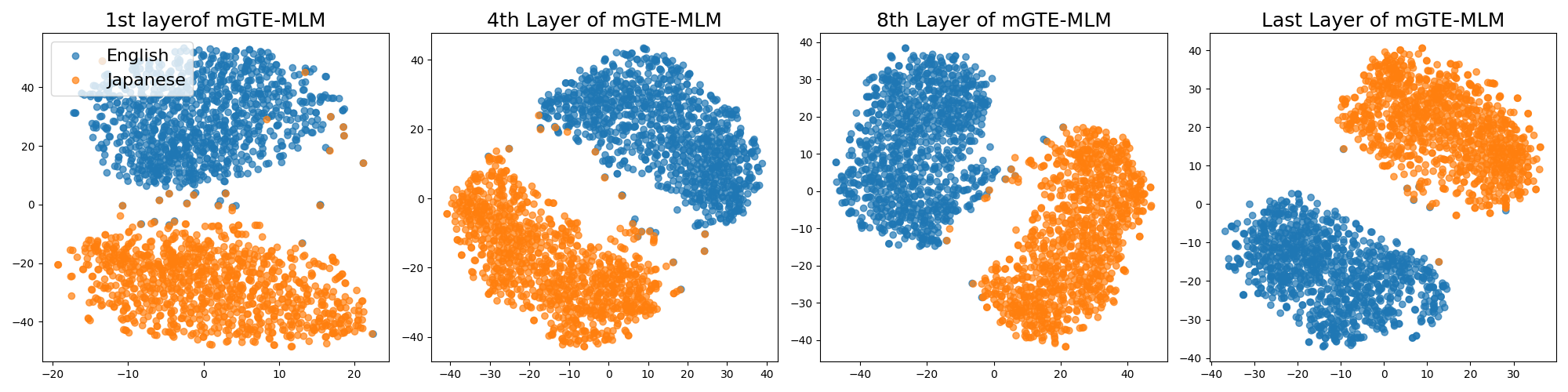} 
    \caption{The t-SNE visualisation of sentence embeddings produced by \textbf{mGTE-MLM-base} (the backbone masked language model of mGTE before fine-tuning) at the 1st, 4th, 8th, and 12th (= last) layers of Transformer for English (blue) and Japanese (orange) sentences. The encoded sentences are 1,000 pairs of translations sampled from a parallel corpus.}
    \label{fig:tsne-gte-mlm-layers}
\end{figure*}

 \begin{figure*}[t!] 
    \centering
    \includegraphics[width=\textwidth]{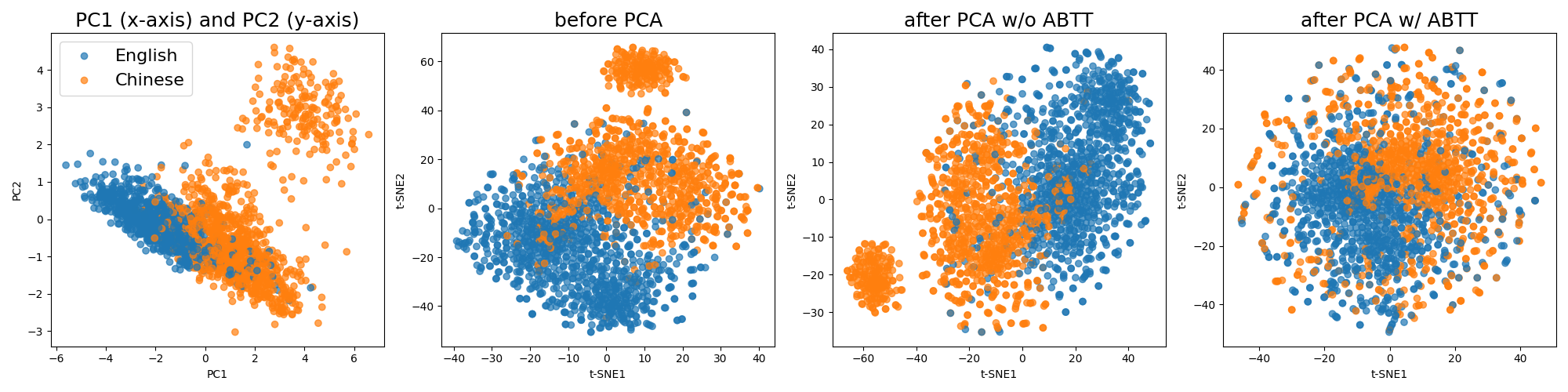} 
    \caption{Scatter plots of our cross-lingual SWEs in \textbf{English (blue) and Chinese (orange)}. The leftmost shows the 1st and 2nd principal components, and the rest are t-SNE visualisation before and after applying PCA with/without ABTT.}
    \label{fig:tsne-enzh}
\end{figure*}

\begin{figure*}[t!] 
    \centering
    \includegraphics[width=\textwidth]{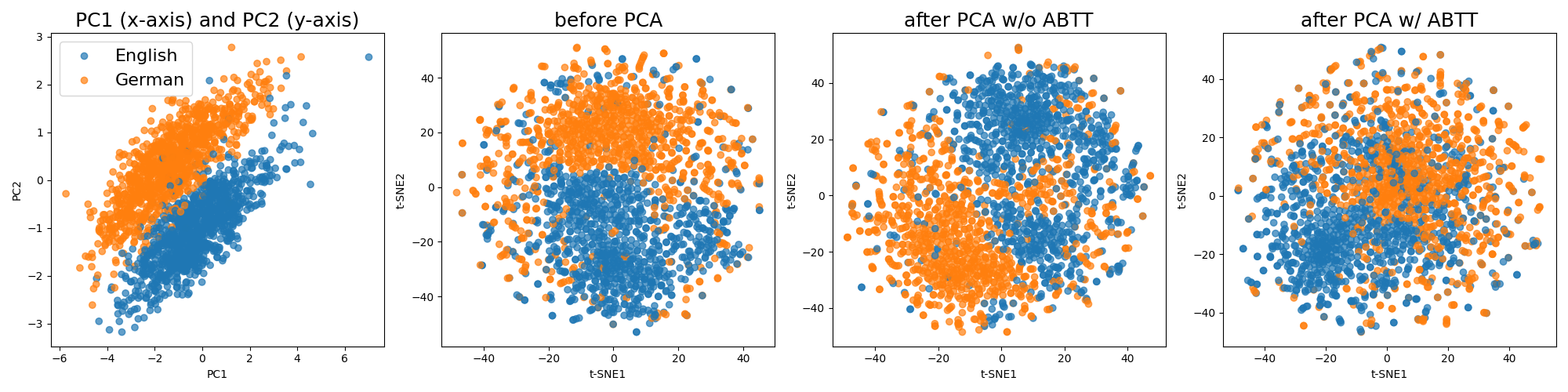} 
    \caption{Scatter plots of our cross-lingual SWEs in \textbf{English (blue) and German (orange)}. The leftmost shows the 1st and 2nd principal components, and the rest are t-SNE visualisation before and after applying PCA with/without ABTT.}
    \label{fig:tsne-ende}
\end{figure*}

\figref{fig:tsne-gte-layers} shows the t-SNE visualisation of sentence embeddings produced by mGTE-base at the 1st, 4th, 8th, and 12th (= last) layers of Transformer. We encode 1,000 translation pairs sampled from an English--Japanese parallel corpus; as such, ideally they should have close representations regardless of language. To obtain sentence vectors, we apply mean pooling at each layer.\footnote{For the last layer, we also tried using the $[$CLS$]$ embedding following the default pooling configuration, and observed a similar result.} The figure shows that the embeddings are completely separated by language at the 1st and 4th layers (just like our SWEs before PCA in \figref{fig:tsne-enja}), but are less clustered at the 8th layer and completely mixed at the the last layer.\footnote{As such, applying PCA did not improve mGTE's performance.} This suggests that mGTE recognises the input language at lower layers, and then captures language-agnostic features such as sentence semantics at higher layers. We also verify this hypothesis by evaluating embeddings at each layer on a translation retrieval task and observing better performance at higher layers.

Notably, this result is somewhat inconsistent with the  previous findings that multilingual masked language models (MLMs) like mBERT have language-specific subspaces in their embedding,  even at near the last layer \cite{gonen-etal-2020-greek,libovicky-etal-2020-language}. One possible reason for this is that mGTE is fine-tuned on various tasks to produce sentence representation, which might make it generate language-agnostic embeddings. To verify this hypothesis, we also evaluate the embeddings of mGTE's backbone MLM (i.e.\ a BERT-like model with several enhancements) before it was fine-tuned on labelled data,\footnote{We use the model at \url{https://huggingface.co/Alibaba-NLP/gte-multilingual-mlm-base}.} and \figref{fig:tsne-gte-mlm-layers} shows the results. It clearly indicates that the embeddings are completely separated based on language at {all} layers, suggesting that mGTE benefits from fine-tuning in producing language-agnostic representations. Before fine-tuning, multilingual MLMs are pre-trained to predict masked tokens in each language at the last layer, and that would naturally encourage them to have language-specific subspaces, as discussed in \citet{gonen-etal-2020-greek}.

\section{Word Embedding Ensembling}\label{sec_ensembling}
Given the success of ensemble learning on many NLP tasks, we try combining $J$ SWEs trained with different algorithms. Specifically, for the input text $z$, we generate its embedding $\mathrm{f}(z)$ as:
\begin{align}
\mathrm{f}(z) &=\frac{1}{\sqrt{\sum_{i=1}^{J} \lambda_i}}\left[\frac{\sqrt{\lambda_i}}{\|\mathrm{f}_i(z)\|}\mathrm{f}_i(z)\right]_{i=1}^J ~,\label{eq_ensemble}\\
\mathrm{f}_i(z)& =\frac{1}{|z|}\sum_{w' \in z}\mathrm{E}_i(w'),
\end{align}
where $\mathrm{f}_i(z) \in \mathbb{R}^{d_i}$ denotes the embedding of $z$ represented by the average of the $i$-th SWE, and $\lambda_i$ is a scalar hyper-parameter that controls the contribution of $\mathrm{f}_i(z)$ to $\mathrm{f}(z)$; $\|\cdot\|$ denotes the vector norm; and $\left[\mathrm{x}_i\right]_{i=1}^J$ denotes the concatenation of the vectors $\mathrm{x}_1, \mathrm{x}_2, \cdots, \mathrm{x}_J$. The dot product of $\mathrm{f}(z)$ exactly corresponds to the weighted average of the cosine similarity measured by each embedding model $\mathrm{f}_i(z)$, with the weight given by $\lambda_i$.  For simplicity, we set $\lambda_1=\dots =\lambda_J=1$ unless otherwise stated.
Notably, unlike ensembling ST models, this method does \textbf{not} increase computational cost very much because we can pre-concatenate the embeddings of $w \in V$ as $\mathrm{{E}}(w) = \left[\mathrm{E}_i(w)\right]_{i=1}^J$, and calculate $\mathrm{f}(z)$ simply by averaging $\mathrm{E}(w)$ and normalising the embedding within each subspace of $\mathrm{E}_i(w)$.\footnote{On the other hand, the cost for calculating similarity and memory usage increase.}

The last three rows in \tabref{result_table} show the results when we ensemble OURS with WordLlama (WL) and Model2Vec (M2V); regarding the last row \textbf{2*OURS + WL + MV}, we set $\lambda_i$ to $2$ for OURS and $1$ for the other models to increase the contribution of OURS. It shows that it performs the best in both Avg-s2s and Avg, indicating the effectiveness of ensembling different SWEs. Looking at the scores on each task, our ensembling method successfully combines the strengths of each model, e.g.\ ensembling OURS with WordLlama is very effective on Reranking and Retrieval, where WordLlama performs the best of all SWEs likely due to its nature of emphasising nominal information (as discussed in Appendix \ref{pos_analysis_detail}).

\begin{table*}[t!]
\begin{center}
\begin{adjustbox}{max width=\textwidth}
\scalebox{0.917}{
\begin{tabular}{lcccccccccc}
\toprule
\multirow{2}{*}{Models  ($d$)} &Class.  &Clust. &PairClass. &Rerank. &Retr. &STS &Summ. &Avg-s2s&Avg\\
\multicolumn{1}{r}{\# Data Sets}&12 &11 &3 &4 &15 &10 &1 &33&56\\\midrule
\multicolumn{10}{c}{Sentence Transformers (STs)}\\\midrule
MiniLM-L6 (384)&63.06&42.35&82.37&58.04&41.95&78.90&30.81&65.95&56.26\\

SimCSE-supervised (768)&67.32&33.43&73.68&47.54&21.82&79.12&\textbf{31.17}&62.75&48.86\\

BGE-base (768)&75.53&45.77&86.55&58.86&53.25&82.40&31.07&71.18&63.55\\

GTE-base (768)&\textbf{77.17}&\textbf{46.82}&85.33&57.66&54.09&81.97&\textbf{31.17}&71.51&64.11\\

LLM2Vec-LLaMa3 (4,096)&{75.92}&{46.45}&\textbf{87.80}&\textbf{59.68}&\textbf{56.63}&\textbf{83.58}&30.94&\textbf{72.94}&\textbf{65.01}\\
\midrule
\multicolumn{10}{c}{{Static Word Embeddings (SWEs)}}\\\midrule

fastText (300)&59.09&30.75&65.77&43.63&22.21&62.82&30.33&53.68&43.05\\
Sent2vec (700)&61.15&31.75&71.88&47.31&27.79&65.80&29.51&56.33&46.29\\

WordLlama (512) [WL] &60.85&36.16&73.72&\textbf{52.82}&\textbf{33.14}&72.47&\textbf{30.83}&60.40&50.23\\

Model2Vec* (512) [M2V] &66.23&35.29&77.89&50.92&32.13&74.22&29.78&62.37&51.33\\

\textbf{OURS} (256) &67.29&36.67&78.99&50.91&31.32&\textbf{75.87}&30.55&63.76&51.97\\
\textbf{OURS} (512) &\textbf{67.96}&\textbf{37.14}&\textbf{79.00}&51.07&32.01&75.59&30.35&\textbf{64.06}&\textbf{52.35}\\

\midrule
\multicolumn{10}{c}{{Ensemble of SWEs}}\\\midrule
\textbf{OURS} + WL (1,024)&66.71&37.56&79.18&\textbf{52.82}&34.13&75.63&30.04&64.10&52.87\\

\textbf{OURS} + WL + MV (1,536)&67.70&37.06&79.37&52.60&\textbf{34.52}&75.63&30.20&64.17&53.09\\

\textbf{2*OURS} + WL + M2V (1,536)&\textbf{68.17}&\textbf{37.60}&\textbf{79.54}&52.39&34.32&\textbf{75.89}&\textbf{30.22}&\textbf{64.55}&\textbf{53.28}\\

\bottomrule
\end{tabular}}
\end{adjustbox}
\end{center}
\caption{Results on MTEB data sets. The last three rows denote the scores when we ensemble OURS with WordLlama [WL] and Model2Vec [M2V] using the method described in Appendix \ref{sec_ensembling}.  The best scores within each subtable are bold-faced. *The scores for Model2Vec in Retr. and Avg are slightly lower than the reported scores, due to a small bug in their MS MARCO evaluation.} 
\label{result_table}
\end{table*}

\begin{table*}[t!]
\begin{center}
\begin{adjustbox}{max width=0.99\textwidth}
\begin{tabular}{llll@{\;}ccc@{\;}ccc@{\;}}
\toprule
\multirow{1}{*}{Models} &\multicolumn{1}{l}{Emb Size $d$} &\multicolumn{1}{l}{Model Path} \\\midrule
\multirow{1}{*}{MiniLM-L6} &384&sentence-transformers/all-MiniLM-L6-v2\\
\multirow{1}{*}{GTE-small} &384&thenlper/gte-small\\
\multirow{1}{*}{GTE-base} &768&Alibaba-NLP/gte-base-en-v1.5\\
\multirow{1}{*}{GTE-large} &1,024&Alibaba-NLP/gte-large-en-v1.5\\
\multirow{1}{*}{mGTE} &768&Alibaba-NLP/gte-multilingual-base\\
\multirow{1}{*}{BGE} &768&BAAI/bge-base-en-v1.5\\

\multirow{1}{*}{SimCSE} &768&princeton-nlp/sup-simcse-bert-base-uncased\\
\multirow{1}{*}{Nomic} &768&nomic-ai/nomic-embed-text-v1.5\\
\multirow{1}{*}{LLaMa3} &4,096&McGill-NLP/LLM2Vec-Meta-Llama-3-8B-Instruct-mntp-supervised\\\midrule
\multirow{1}{*}{fastText} &300&crawl-300d-2M.vec\\
\multirow{1}{*}{Sent2vec} &700&sent2vec\_wiki\_bigrams\\
\multirow{1}{*}{WordLlama} &512&wordllama-l3-supercat\\
\multirow{1}{*}{Model2Vec} (English) &512&potion-base-32M\\
\multirow{1}{*}{Model2Vec (cross-lingual)} &256&M2V\_multilingual\_output\\

\bottomrule
\end{tabular}
\end{adjustbox}
\end{center}
\caption{Details of the models used in our experiments.}
\label{model_details}
\end{table*}

\section{Model Details}\label{model_detail}

\tabref{model_details} lists the details of the models used in our paper. The Sent2vec model is downloaded from  
\url{https://github.com/epfml/sent2vec}; fastText from \url{https://fasttext.cc/docs/en/english-vectors.html} \cite{mikolov-etal-2018-advances}; and the rest from Hugging Face (\url{https://huggingface.co/models}).

\begin{table*}[t!]
\begin{center}
\begin{adjustbox}{max width=\textwidth}
\scalebox{1}{
\begin{tabular}{lcccccccccc}
\toprule
\multirow{2}{*}{Models  ($d$)} &Class.  &Clust. &PairClass. &Rerank. &Retr. &STS &Summ. &Avg-s2s&Avg\\
\multicolumn{1}{r}{\# Data Sets}&12 &11 &3 &4 &15 &10 &1 &33&56\\\midrule
\multicolumn{10}{c}{Sentence Transformers (STs)}\\\midrule
SimCSE-supervised (768)&67.32&33.43&73.68&47.54&21.82&79.12&\textbf{31.17}&62.75&48.86\\
Nomic (768)&	
73.55&43.93&84.61&55.78&53.01&81.94&30.40&69.52&62.28\\
GTE-small (384)&72.31&44.89&	
83.54&57.70&49.46&\textbf{82.07}&30.42&69.32&61.36\\

GTE-base (768)&{77.17}&{46.82}&\textbf{85.33}&57.66&54.09&{81.97}&\textbf{31.17}&71.51&64.11\\
GTE-large (1,024)&\textbf{77.75}&\textbf{47.96}&84.53&\textbf{58.50}&\textbf{57.91}&81.43&	
30.91&\textbf{71.65}&\textbf{65.39}\\
\midrule
\multicolumn{10}{c}{Static Word Embeddings (SWEs)}\\\midrule
OURS-SimCSE (256)&64.72&32.78&75.65&48.36&26.54&74.83&30.61&61.16&48.83\\
OURS-Nomic (256)&65.94&34.74&\textbf{79.29}&50.15&30.47&75.49&\textbf{31.49}&62.69&50.99\\
OURS-GTE-small (256)&63.90&34.14&78.17&49.66&27.39&72.63&31.14&60.76&49.00\\
OURS-GTE-base (256) &\textbf{67.29}&\textbf{36.67}&78.99&\textbf{50.91}&31.32&\textbf{75.87}&30.55&\textbf{63.76}&\textbf{51.97}\\
OURS-GTE-large (256)&65.36&36.11&78.13&50.42&\textbf{31.58}&75.59&30.73&62.91&51.39\\

\bottomrule
\end{tabular}}
\end{adjustbox}
\end{center}
\caption{Performance of different ST models and our models ($d=256$) distilled from each ST.} \label{tab_other_model}
\end{table*}

\begin{table*}[t!]
\begin{center}
\begin{adjustbox}{max width=\textwidth}
\scalebox{1}{
\begin{tabular}{lcccccccccc}
\toprule
\multirow{2}{*}{Models  ($d$)} &Class.  &Clust. &PairClass. &Rerank. &Retr. &STS &Summ. &Avg-s2s&Avg\\
\multicolumn{1}{r}{\# Data Sets}&12 &11 &3 &4 &15 &10 &1 &33&56\\\midrule
\textbf{OURS} ($d$ = 64,~~ $\vert V \vert$ = 150k) &60.20&29.09&60.12&44.07&17.83&62.12&30.07&52.51&41.39\\
\textbf{OURS} ($d$ = 128, $\vert V \vert$ = 150k) &65.92&35.51&78.05&50.09&28.33&75.06&\textbf{30.60}&62.62&50.40\\
\textbf{OURS} ($d$ = 256, $\vert V \vert$ = 150k) &67.29&36.67&78.99&50.91&31.32&\textbf{75.87}&30.55&63.76&51.97\\
\textbf{OURS} ($d$ = 512, $\vert V \vert$ = 150k) &\textbf{67.96}&\textbf{37.14}&\textbf{79.00}&\textbf{51.07}&\textbf{32.01}&75.59&30.35&\textbf{64.06}&\textbf{52.35}\\\midrule
\textbf{OURS} ($d$ = 256, $\vert V \vert$ = 150k) &67.29&\textbf{36.67}&\textbf{78.99}&\textbf{50.91}&\textbf{31.32}&{75.87}&30.55&\textbf{63.76}&\textbf{51.97}\\

\textbf{OURS} ($d$ = 256, $\vert V \vert$ = 100k) &67.29&36.12&78.56&50.71&31.18&\textbf{75.90}&\textbf{30.84}&63.55&51.80\\
\textbf{OURS} ($d$ = 256, $\vert V \vert$ = 75k) &\textbf{67.38}&35.67&78.55&50.59&30.56&75.67&30.36&63.36&51.51\\
\textbf{OURS} ($d$ = 256, $\vert V \vert$ = 50k) &67.26&34.11&78.19&50.30&30.25&75.21&30.68&62.70&50.97\\
\textbf{OURS} ($d$ = 256, $\vert V \vert$ = 30k) &66.90&32.89&78.03&50.00&28.61&75.14&29.81&62.25&50.16\\

\bottomrule
\end{tabular}}
\end{adjustbox}
\end{center}
\caption{Performance of OURS trained with different embedding and vocabulary sizes.} \label{detail_model_size}

\end{table*}

\end{document}